\documentclass[letterpaper]{article} 
\usepackage{aaai2026}  
\usepackage{times}  
\usepackage{helvet}  
\usepackage{courier}  
\usepackage[hyphens]{url}  
\usepackage{graphicx} 
\usepackage{natbib}  
\usepackage{caption} 
\usepackage{algorithm}
\usepackage{algorithmic}

\usepackage{newfloat}
\usepackage{listings}
\DeclareCaptionStyle{ruled}{labelfont=normalfont,labelsep=colon,strut=off} 
\floatstyle{ruled}
\newfloat{listing}{tb}{lst}{}
\floatname{listing}{Listing}
\usepackage{url}            
\usepackage{booktabs}       
\usepackage{amsfonts}       
\usepackage{nicefrac}       
\usepackage{microtype}      
\usepackage{xcolor}
\usepackage{multirow}

\usepackage{color}

\setcitestyle{numbers,square}
\title{Hypothesis Generation via LLM-Automated Language Bias for ILP}

\author{
    Yang Yang\textsuperscript{\rm 1}\thanks{Equal Contribution}, 
    Jiemin Wu\textsuperscript{\rm 1*}, 
    Yutao Yue\textsuperscript{\rm 1}\thanks{Corresponding Author}, 
}
\affiliations{
    \textsuperscript{\rm 1}The Hong Kong University of Science and Technology (Guangzhou) \\
    \{yyang945, jwu663\}@connect.hkust-gz.edu.cn \\
    yutaoyue@hkust-gz.edu.cn
    
}

\usepackage{amsmath}
\usepackage{listings}
\usepackage{xcolor}

\begin{document}
\maketitle
\begin{abstract}
Inductive Logic Programming (ILP) is a principled approach for generalizing regularities from data and constructing hypotheses as interpretable logic programs. However, a key limitation is its reliance on expert-crafted language bias—the predicate inventory, types, and mode declarations that delimit the search space. We propose hypothesis generation via LLM-automated language bias: multi-agent LLMs design the bias from raw text and translate descriptions into typed facts, and a robust ILP solver induces rules under a global consistency objective. This approach reduces traditional ILP’s reliance on predefined symbolic structures and the noise sensitivity of LLM-only pipelines that directly generate hypotheses as text or code. Extensive experiments in diverse, challenging scenarios validate superior performance, providing a practical, explainable, and verifiable route to hypothesis generation.

\end{abstract}

\section{Introduction}

Hypothesis generation—the process of forming systematic explanations from fragmented observations and iteratively validating them—plays a central role in advancing artificial intelligence \citep{lake2015human}. This "generate-test" paradigm is fundamental not only in scientific discovery but also in practical AI tasks, such as diagnosing the root cause of software defects from error logs \citep{zeller2009programs}. The capacity to continuously produce and verify hypotheses is thus crucial for building robust AI systems, especially when deployed in open environments or high-stakes domains like medical diagnosis and financial decision-making \citep{letham2015interpretable,rudin2019stop}.

Inductive Logic Programming (ILP), a traditional method for hypothesis generation, discovers knowledge by searching rule sets within expert-defined predicate spaces \citep{cropper2022inductive}. This reliance on expert-crafted predicates, however, poses significant challenges in complex domains. For instance, in areas like protein interaction research, specialists must meticulously predefine predicates capturing crucial domain-specific features (e.g., atomic distances or amino acid properties) and manually encode experimental data into symbolic facts to enable the derivation of reliable binding site identification rules \citep{a2012automated}.Meanwhile, current ILP research has focused primarily on optimizing rule search algorithms\citep{cropper2021learning}, rarely exploring the automatic construction of predicate spaces, which severely limits its scalability in open-domain tasks.

The emergence of Large Language Models (LLMs) offers a new pathway for hypothesis generation. Their end-to-end learning paradigm can directly produce candidate hypotheses from unstructured texts (e.g., equipment failure logs or experimental reports) in various forms, including natural language \citep{zhou2024hypothesis}, logical expressions \citep{luo2023chatrule}, or even code \citep{qiu2023phenomenal}. This openness bypasses the traditional reliance on expert-crafted predicates. However, LLMs still face critical challenges: firstly, significant sensitivity to noise—for instance, hypothesis accuracy can plummet from 71.2\% to 50.9\% with a 12.5\% increase in input data noise \citep{qiu2023phenomenal}. Secondly, while an LLM might generate numerous plausible individual hypotheses \citep{zhou2024hypothesis}, its typically heuristic generation process makes it difficult to assemble these into a compact, internally consistent, and collectively optimal set of rules required to comprehensively describe complex phenomena—a stark contrast to ILP systems that often learn such interdependent rule programs \citep{nickles2015hybrid,manhaeve2021neural}. 

To harness the semantic strengths of LLMs while retaining the rigorous, verifiable outputs of ILP, we introduce a collaborative reasoning framework, shown in Figure \ref{fig:figure1}. This framework first employs a multi-agent LLM system to automate the generation of a structured language bias, particularly the predicate system, directly from raw text. Subsequently, this LLM-generated bias guides the transformation of large-scale textual data into symbolic facts. This structured knowledge then empowers an ILP engine to conduct a robust, constrained search, yielding a globally coherent and optimal set of rules. 
Unlike prior approaches that evaluate under idealized conditions (e.g., zero label noise or fixed few-shot claims such as “8 examples suffice”~\citep{qiu2023phenomenal}), we conduct a controlled robustness study along four axes: (i) surface-template perturbations (paraphrases, clause/order shuffling, light numeric fuzzing), (ii) label noise with flip rates $p!\in!{0.05,0.10,0.20}$, (iii) class imbalance with positive:negative ratios up to $1!:!9$, and (iv) relational complexity (unary \textit{SHOES} vs. relational \textit{ZENDO}). We further assess model-agnostic stability by swapping LLM backends and reporting cross-model variance, substantiating our method's superior efficacy and robustness.

While related explorations combine LLMs with symbolic reasoning, notably for formal verification \citep{olausson2023linc,kalyanpur2024llm,jiang2024leanreasoner}, they often employ LLMs as mere translators to \emph{predefined} logical forms. Our work, in contrast, pioneers LLM-driven \emph{automated symbolic template generation}. This process dynamically creates the entire logical scaffolding that constitutes the Inductive Logic Programming (ILP) \emph{language bias}—including the core predicate system defining concepts and relations. Traditionally, defining this bias to guide ILP's search and ensure its effectiveness requires extensive expert input. Automating the creation of this guiding structure thus unlocks ILP's potential to dynamically adapt to novel problem domains and discover verifiable hypotheses where manual bias definition was previously a prohibitive barrier.

The main contributions of this study are as follows:

\noindent \textbf{1)} We introduce a novel multi-agent framework using LLMs to automate ILP language bias (predicate system) construction. This pioneers an end-to-end pipeline from unstructured text to verifiable hypotheses, advancing explainable hybrid AI.

\noindent \textbf{2)} Unlike prior work limited to idealized data, we systematically evaluate LLM-based induction across challenging data dimensions (e.g., noise, imbalance, complexity), enabling a more thorough and realistic capability assessment.

\noindent \textbf{3)} Extensive experiments demonstrate our framework's superior accuracy, robustness against data perturbations, and generalization across LLMs, significantly outperforming existing baselines.

\section{Preliminaries}
In First-Order Logic (FOL), \emph{predicates} are used to describe objects or the relationships between objects and can be classified according to the number of arguments, such as unary or binary. For example, the unary predicate \verb|isRed(x)| denotes "x is red," and the binary predicate \verb|parent(x, y)| denotes "x is a parent of y." Instantiating the arguments of a predicate to constants (e.g., \verb|parent(Alice, Bob)|) yields an \emph{atom}; if this atom is considered to be true, it is called a \emph{fact}. Typically, the known information in a domain consists of several facts, which are regarded as directly usable \emph{background knowledge}.

Building on this foundation, more general inference rules can be expressed using a \emph{Horn clause}, which is typically written in the form \(H \;\leftarrow\; B_1 \;\wedge\; B_2 \;\wedge\; \dots \;\wedge\; B_k\), where \(H\) is the rule's "head" and each \(B_i\) constitutes the rule's "body". Its semantics is that if all atoms \(B_i\) in the body are true, then the head \(H\) must also be true. For example, to express "if \(x\) is a parent of \(y\) and \(y\) is a parent of \(z\), then \(x\) is an ancestor of \(z\)", one can write \( \verb|ancestor(x, z)| \;\leftarrow\; \verb|parent(x, y)| \;\wedge\; \verb|parent(y, z)|\). Here, \(\verb|ancestor(x, z)|\) serves as the rule's head, while \(\verb|parent(x, y)|\) and \(\verb|parent(y, z)|\) form the rule's body. Multiple such Horn clauses can be assembled into a \emph{rule set}, typically exhibiting an "OR-of-ANDs" structure. In such a set, if all preconditions (the body) of any individual rule are satisfied (based on background knowledge), its conclusion (the head) is considered true.

\section{Related Work}
\subsection{Inductive Logic Programming}
Inductive Logic Programming (ILP) automatically learns interpretable logic programs from positive and negative examples of a target predicate, along with background knowledge, by searching for rule sets in the First-Order Logic space. Broadly, these approaches can be categorized into heuristic methods (e.g., FOIL \citep{quinlan1990learning}, Progol \citep{muggleton1995inverse}, Aleph \citep{srinivasan2001aleph}), constraint-solving techniques (e.g., ILASP \citep{law2014inductive}, Popper \citep{cropper2021learning}, MAXSYNTH \citep{hocquette2024learning}), and differentiable approaches (e.g., \citep{evans2018learning,glanois2022neuro,sen2022neuro}). However, all these approaches are essentially search algorithms that depend on an expert-defined \emph{language bias} to define the search space. This language bias consists of the set of permissible predicates, the structural forms that rules can take, and other constraints that collectively restrict the universe of possible hypotheses the system can consider.

Unlike prior research focused on refining search within fixed language biases, our work pioneers LLM-driven automation of the \emph{language bias} itself. This encompasses formulating the predicate system, key structural constraints, and other declarative elements, all traditionally demanding extensive expert input. Our primary objective is to eliminate manual language bias engineering, thus enabling more adaptive integration and broader applicability of established ILP algorithms, particularly in open-domain tasks.

\subsection{Hypothesis Generation Based on LLMs}
Large Language Models (LLMs) have recently garnered significant attention for hypothesis generation. For instance, ChatRule \citep{luo2023chatrule} employs LLMs to directly derive logical rules from knowledge graphs for explainable reasoning, while Moose-Chem \citep{yang2024moose} utilizes multi-turn inspiration selection and evolutionary algorithms to propose novel molecular hypotheses. Other notable approaches include HypoGeniC \citep{zhou2024hypothesis}, which uses a multi-armed bandit-like mechanism for iterative rule generation and filtering; Iterative Hypothesis Refinement \citep{qiu2023phenomenal}, guiding LLMs through a “propose-select-refine” process for concept-level rule abstraction; and HtT \citep{zhu2023large}, which compiles LLM-generated candidate rules into an executable library for inference.

While generalizable, LLMs' noise sensitivity and heuristic hypothesis generation often yield sub-optimal or incoherent rule sets. Our approach counters this: LLMs first auto-construct a \emph{predicate system} from unstructured data, which then allows Inductive Logic Programming (ILP) methods using precise constrained solving to produce a globally coherent and optimal rule set. This hybrid methodology yields flexible, robust, and interpretable hypotheses, particularly effective in complex, noisy, and open-domain settings.

\section{Methodology}

\begin{figure*}[t]
    \centering
    \includegraphics[width=0.8\textwidth]{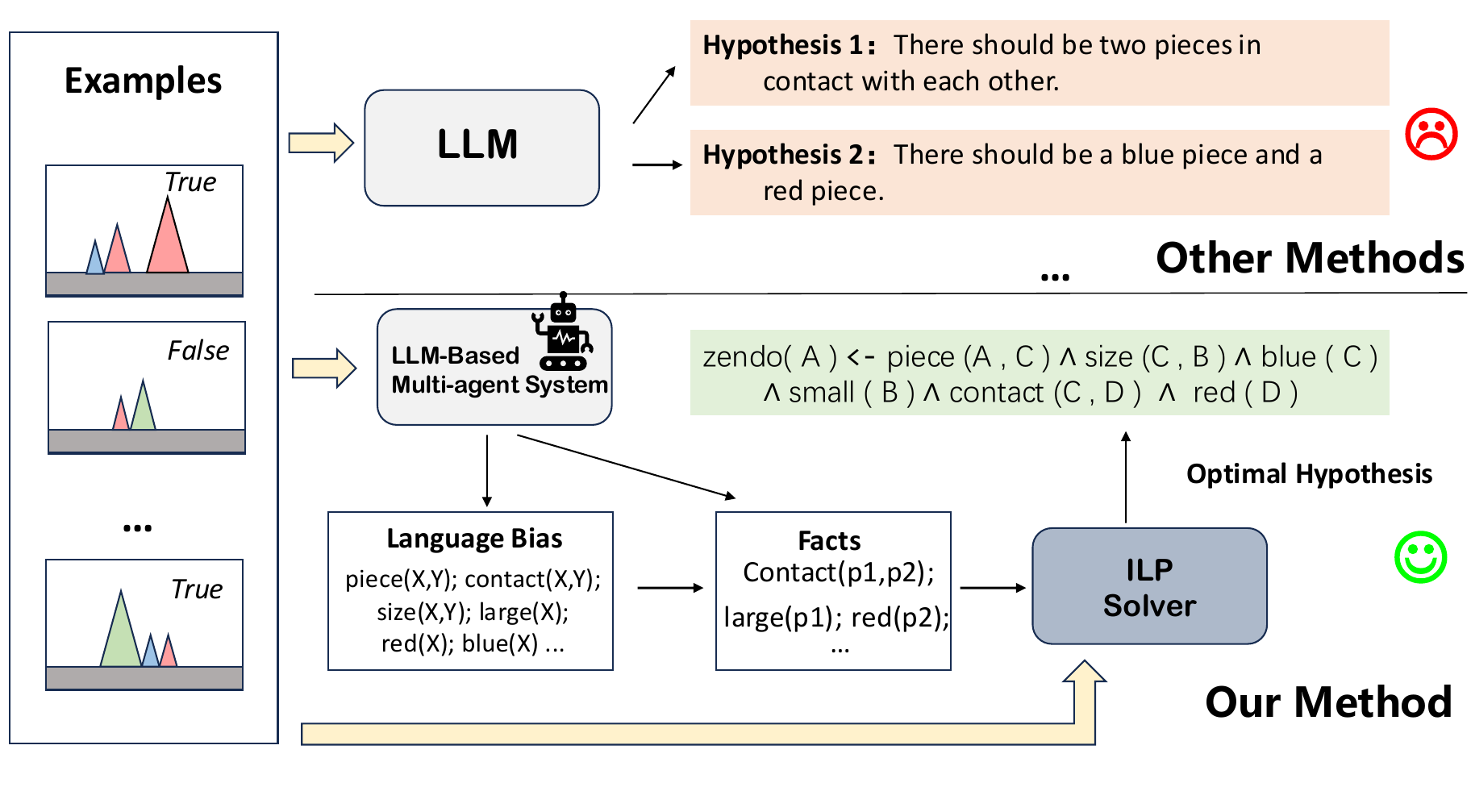}
  \caption{\textbf{Illustration of our pipeline and example rule.}
  The LLMs produce an ILP \emph{language bias} and \emph{typed facts}, and a verifiable ILP solver induces rules under a global-consistency objective.
  \textbf{Example hypothesis returned by the solver:}
  \texttt{zendo(A) :- piece(A,C), size(C,B), blue(C), small(B), contact(C,D), red(D).}
  \textbf{Meaning (variable types):} $A$: scene, $C,D$: pieces, $B$: size attribute.
  The rule states that $zendo(A)$ holds \emph{iff} there exists a \emph{blue} piece $C$ in scene $A$ that is \emph{small} (via $size(C,B)\,\wedge\,small(B)$) and is \emph{in contact} with a \emph{red} piece $D$.}
    \label{fig:figure1}
    \vspace{-1em}
\end{figure*}

Our approach contain three core stages: Predicate System Construction, Symbolic Knowledge Encoding, and ILP Learning. The overall framework is shown in Figure \ref{fig:figure1}.


\subsection{Predicate System Construction}
\label{subsec:predicate-construction}
The construction of the predicate system is a cornerstone of our framework, directly influencing the quality and efficiency of subsequent symbolic encoding and Inductive Logic Programming (ILP) learning. This process is driven by a multi-agent subsystem, primarily comprising an Actor agent and a Critic agent. The Actor is responsible for initially designing and iteratively refining the predicate system based on raw text samples, while the Critic meticulously evaluates the Actor's proposals and provides guiding feedback. Through multiple rounds of collaborative interaction between the Actor and Critic, the system automatically generates a predicate system that is highly relevant to the task, structurally sound, and compliant with the requirements of an ILP solver.

\paragraph{Actor Agent}
The Actor's role is to design and optimize the predicate system, either from scratch or based on feedback from the previous iteration. It receives a small subset of training samples and is guided by a few-shot examples of predicate abstraction from other general tasks, alongside predefined predicate design principles and constraints. Based on these inputs, the Actor generates a complete definition of the predicate system, encompassing:
\noindent \textbf{Core Predicates:} Typically includes one target Head Predicate, representing the core concept to be learned and predicted (e.g., \texttt{suitable\_for\_business/1}), and multiple Body Predicates describing sample attributes and relationships (e.g., \texttt{formal\_shoes/1}, \texttt{leather/1}, etc.). The number of body predicates is flexible and determined by task requirements.
\noindent \textbf{Predicate Formalism:} For each predicate, its arity (number of arguments) and the type of each argument are explicitly defined (e.g., \texttt{type(formal\_shoes, (shoes,))}).
\noindent \textbf{Declarations and Constraints:} Other meta-information required by the ILP solver is also defined, such as the input/output modes for predicate arguments (e.g., \texttt{direction(formal\_shoes, (in,))}) and global constraints like the maximum number of variables (\texttt{max\_vars}) or body length (\texttt{max\_body}) in a clause.
The Actor outputs the generated predicate system in a textual format (typically Prolog-compatible) for evaluation by the Critic.

\paragraph{Critic Agent}
The Critic is responsible for comprehensively evaluating the predicate system generated by the Actor, primarily from semantic and syntactic perspectives. Its evaluation includes:
\noindent \textbf{Semantic Evaluation:} The Critic leverages the understanding capabilities of Large Language Models (LLMs) to analyze the semantic appropriateness of the predicate system. This includes checking for completeness (i.e., whether crucial concepts are missed), redundancy (i.e., whether semantically similar predicates exist), and relevance to the target task.
\noindent \textbf{Syntactic and Constraint Validation:} The Critic also validates the predicate system against predefined syntactic rules and constraints using programmatic logic, ensuring its structural compliance with ILP solver specifications. For instance, it verifies that each argument type in the head predicate is covered by at least one body predicate and checks the correctness of arity and type declarations.

Based on these checks, the Critic generates an evaluation summary and determines if the current predicate system is satisfactory. If not, the identified issues in the summary are fed back to the Actor for the next round of refinement. The predicate system is finalized and used for subsequent symbolic knowledge encoding and ILP learning only when it passes all checks or when a predefined maximum number of iterations (set to five in our experiments) is reached.

\subsection{Symbolic Knowledge Encoding}
\label{subsec:data-transformation}
Following predicate system finalization (Section \ref{subsec:predicate-construction}), our \textbf{Translator} agent transforms natural language samples into Prolog facts. It parses each sample, mapping textual features to the established predicates. This translation proceeds in batches, circumventing LLM context limitations and the need for simultaneous full-dataset access common in other approaches. To bolster stability, translation failures trigger a retry (max two attempts). This systematic, batch-oriented encoding bridges natural language semantics with formal logic, enabling scalable conversion to a symbolic representation for ILP.

For example, consider the SHOES dataset. The LLM-derived predicate system might define the head predicate as \texttt{type(suitable\_for\_business,(shoes,)).} and various body predicates such as \texttt{type(formal\_shoes,(shoes,)).}, \texttt{type(black,(shoes,)).}, \texttt{type(leather,(shoes,)).}, \texttt{type(expensive,(shoes,)).}, etc. Given a textual sample like ``\textit{Shoe\_001 is a black formal shoe made of leather, expensive in price and very comfortable to wear. This shoe is suitable for business,}'', the symbolic knowledge encoding module would process this to generate Prolog facts. The background knowledge facts might be: \texttt{black(shoe\_001).}, \texttt{formal\_shoes(shoe\_001).}, \texttt{leather(shoe\_001).}, \texttt{expensive(shoe\_001).}, \texttt{very\_comfortable(shoe\_001).} The example fact indicating the target classification would be: \texttt{pos(suitable\_for\_business(shoe\_001)).} This structured conversion ensures that all relevant information from the text is translated into a consistent logical format, forming the empirical basis for the subsequent ILP learning stage.

\subsection{ILP Learning}
Upon completion of the symbolic knowledge encoding (Section \ref{subsec:data-transformation}), the comprehensive set of structured Prolog facts, along with the LLM-generated predicate system (Section \ref{subsec:predicate-construction}), is provided as input to an ILP solver. 
We employ MAXSYNTH \citep{hocquette2024learning}, which is an advanced solver that applies the Minimum Description Length (MDL) principle to balance rule complexity with noise coverage. This enables it to find globally optimal or near-optimal rule sets even with imperfect data, achieving both search efficiency and robustness, particularly in the presence of label noise.
To enhance stability, if ILP solving fails, the algorithm restarts from predicate system design, with up to two attempts.

When successful, the ILP solver outputs a set of Horn clauses as the final learned hypothesis. These are directly interpretable logical formulas explaining the target Head Predicate. For instance, on the SHOES dataset, a learned rule set might include: Rule 1: \( \verb|suitable_for_business(A)| \leftarrow \verb|expensive(A)| \wedge \verb|formal_shoes(A)| \); and Rule 2: \( \verb|suitable_for_business(A)| \leftarrow \verb|synthetic_leather(A)| \wedge \verb|very_comfortable(A)| \).
This integration of an advanced ILP solver enhances the applicability and robustness of rule-based hypothesis generation in open-domain tasks.

\begin{table*}[t]
  \centering
  \footnotesize
    \begin{tabular}{c|l|rr|rr|rr}
    \toprule
    \midrule
    \multirow{2}[4]{*}{Method} & \multicolumn{1}{c|}{\multirow{2}[4]{*}{Model}} & \multicolumn{2}{c|}{Shoes} & \multicolumn{2}{c|}{Zendo} & \multicolumn{2}{c}{Average} \\
\cmidrule{3-8}          &       & \multicolumn{1}{l}{Acc} & \multicolumn{1}{l|}{F1} & \multicolumn{1}{l}{Acc} & \multicolumn{1}{l|}{F1} & \multicolumn{1}{l}{Acc} & \multicolumn{1}{l}{F1} \\
    \midrule
    \multirow{3}[2]{*}{Iterative Hypothesis Refinement} 
          & GPT-4o            &   96.7   &   96.7   &   50.0   &    34.1   &   73.4   &   65.4\\
          & Claude-3.7-sonnet &   98.3   &   98.3   &   60.0   &    45.1   &   79.2   &   71.7\\
          & DeepSeek-V3       &   95.0   &   95.0   &   46.7   &    30.4   &   70.9   &   62.7\\
          & Qwen3-32B         &   -   &   -   &   -   &    -   &   -   &   -\\
    \midrule
    \multirow{3}[2]{*}{HypoGeniC} 
          & GPT-4o            &   51.7    &   49.9    &   73.3    &   71.8    &   62.5  &  60.85\\
          & Claude-3.7-sonnet &   75.0    &   74.3    &   68.3    &   66.6    &   71.65 &  70.45\\
          & DeepSeek-V3       &   70.0    &   67.9    &   70.0    &   69.1    &   70.0  &  68.5\\
          & Qwen3-32B         &   83.3    &   82.0    &   46.7    &   37.2    &   65.0  &  59.6\\
    \midrule
    \multirow{3}[2]{*}{Ours} 
          & GPT-4o            &   87.9    &   87.9    &   76.7    &    75.4   &   82.3  &  81.7\\
          & Claude-3.7-sonnet &   88.3    &   88.1    &   81.3    &    81.4   &   84.8  &  84.8\\
          & DeepSeek-V3       &   88.3    &   88.1    &   81.3    &    81.4   &   84.8  &  84.8\\
          & Qwen3-32B         &   87.9    &   87.9    &   80.0    &    80.7   &   84.0  &  84.3\\
    \midrule
    \bottomrule
    \end{tabular}%

    \caption{
    Comparison of hypothesis generation performance on the Shoes and Zendo datasets. Accuracy (Acc, \%) and F1 score (F1, \%) are reported for each dataset and their average.
    }
  \label{tab:main_res}
\end{table*}

\section{Experiment Setup}
\subsection{Datasets, Baselines and Metrics}
\textbf{Datasets:}
We consider two synthetic binary classification tasks: \textit{SHOES} and \textit{ZENDO}. \textit{SHOES} is constructed by us to evaluate models' ability to determine the suitability of shoes for business occasions, with all attributes expressible as unary predicates (e.g., \texttt{Black(X)}, \texttt{leather(X)}). \textit{ZENDO} is adapted from classic cognitive psychology experiments and is more challenging: it also involves binary predicates (e.g., \texttt{contact(X, Y)}, \texttt{has\_piece(X, Y)}), requiring models to reason about more complex logical relations (see Appendix for details). For each task, we first specify a set of rules and generate corresponding logical facts to construct samples, which are then further converted into natural language form using templates.

\textbf{Baselines:}
We consider two LLM-based inductive reasoning algorithms as baselines: \textit{HypoGeniC}\citep{zhou2024hypothesis} and \textit{Iterative Hypothesis Refinement (IHR)}\citep{qiu2023phenomenal}. \textit{HypoGeniC} generates hypotheses in natural language form and iteratively improves them using a bank of counterexamples. \textit{Iterative Hypothesis Refinement} further enhances this process by generating, selecting, and refining hypotheses, and is also capable of producing executable code as candidate rules (see Appendix for details).

\textbf{Metrics.} We assess the reasonableness of a hypothesis by treating it as a decision results, using prediction \textbf{Accuracy} and \textbf{Macro-F1}.For text-outputting methods (HypoGeniC), following its original protocol we present the textual hypothesis plus the new example to the same LLM to obtain prediction label.

\subsection{Experimental Variables}

To systematically evaluate LLMs' hypothesis generation capabilities under diverse data conditions, we consider the following dataset-level variables:

\noindent \textbf{Rule Num} refers to the number of underlying logical rules guiding the generation of facts for each sample. For example, a Zendo dataset with two rules may require "a red object on the left" (rule 1) or "a green object adjacent to a blue object" (rule 2), with positive samples satisfying either rule. Our experiments include rule sets containing 1, 2, or 3 rules.

\noindent \textbf{Template Num} refers to the number of natural language templates used to describe the same logical fact. Each sample is randomly expressed using one of several candidate templates, potentially increasing difficulty for LLMs to abstract consistent semantics across diverse expressions. We evaluate conditions with 1, 2, or 3 candidate templates.

\noindent \textbf{Sample Size} refers to the total number of samples in the dataset. With smaller sample sizes, LLMs may struggle to capture underlying patterns, limiting generalization capabilities. We compare performance with 50, 100, and 200 samples.

\noindent \textbf{Positive Ratio} refers to the proportion of positive samples in the dataset. Lower positive ratios may cause models to overlook minority classes, resulting in poor coverage of positive cases. Our experiments include positive ratios of 20\%, 30\%, and 50\%.

\noindent \textbf{Noise Ratio} refers to the probability of randomly flipped labels in the training set (test labels remain noise-free). Higher noise levels can mislead models, decreasing accuracy and stability. We evaluate noise levels of 0\%, 10\%, and 20\%.

\subsection{Implementation Details}  
To further evaluate the generality of each method, we consider the following language models in our experiments: GPT-4o \citep{gpt4o}, Claude-3.7-sonnet \citep{anthropic_claude_2025}, DeepSeek-V3 \citep{liu2024deepseek}, and Qwen3-32b \citep{yang2024qwen2}. The temperature parameter is set to 0 to reduce generation randomness. The dataset is split into 80\% for training and 20\% for testing. For each experiment, we perform three independent dataset generation processes and report the average results on the test set across the three runs.

\section{Experiments and Results}
Based on the experimental setup, we evaluate our method by addressing the following research questions:

\textbf{RQ1:} How sensitive is our method to the choice of LLMs, and does it show generality?

\textbf{RQ2:} Can our method maintain stable performance under various data scenarios?

\textbf{RQ3:} How does the overall performance of our method compare to existing advanced baselines?

\subsection{Main Experiments}
In the main experiments, we fix five data generation variables for fair comparison: rule number (2), noise ratio (10\%), template number (2), sample size (100), and positive ratio (50\%). Each method is evaluated using four mainstream LLMs (GPT-4o, Claude-3.7-sonnet, DeepSeek-V3, and Qwen3-32b) to systematically assess performance differences across models.

\subsubsection{Comparative Analysis of Methods}
Table \ref{tab:main_res} presents results across datasets and models. Our method demonstrates superior performance, particularly on the complex ZENDO task. HypoGeniC shows adaptability but lacks consistency, with performance heavily dependent on the underlying LLM capabilities. IHR excels on simple tasks but struggles with complex reasoning challenges and fails completely with Qwen3-32B. Our method's key advantage is its model-agnostic design, maintaining consistent performance by delegating logical reasoning to an ILP solver while using LLMs for language understanding.

\subsubsection{Dataset Complexity Analysis}
The results across datasets reveal important differences in method capabilities. IHR achieves near-perfect performance on the simpler SHOES task (96-98\% accuracy) but experiences a dramatic drop of almost 50\% on the more complex ZENDO task, indicating difficulties with relational reasoning involving binary predicates. Our method maintains more consistent performance across both datasets, demonstrating robust reasoning capabilities regardless of task complexity.

\subsubsection{Model Dependency Analysis}
The results show significant differences in model dependency across methods. HypoGeniC exhibits high variability, with performance differences exceeding 30\% between models on certain tasks. IHR shows considerable model dependency, especially for complex reasoning. In contrast, our method demonstrates remarkable consistency across all four LLMs, with performance variations typically below 5\%. Claude-3.7-sonnet and DeepSeek-V3 achieve identical performance with our approach, while GPT-4o and Qwen3-32B show only minor differences, highlighting our method's practical advantage for deployment across different environments.

\begin{figure*}[!t]
    \centering
    \includegraphics[width=\textwidth]{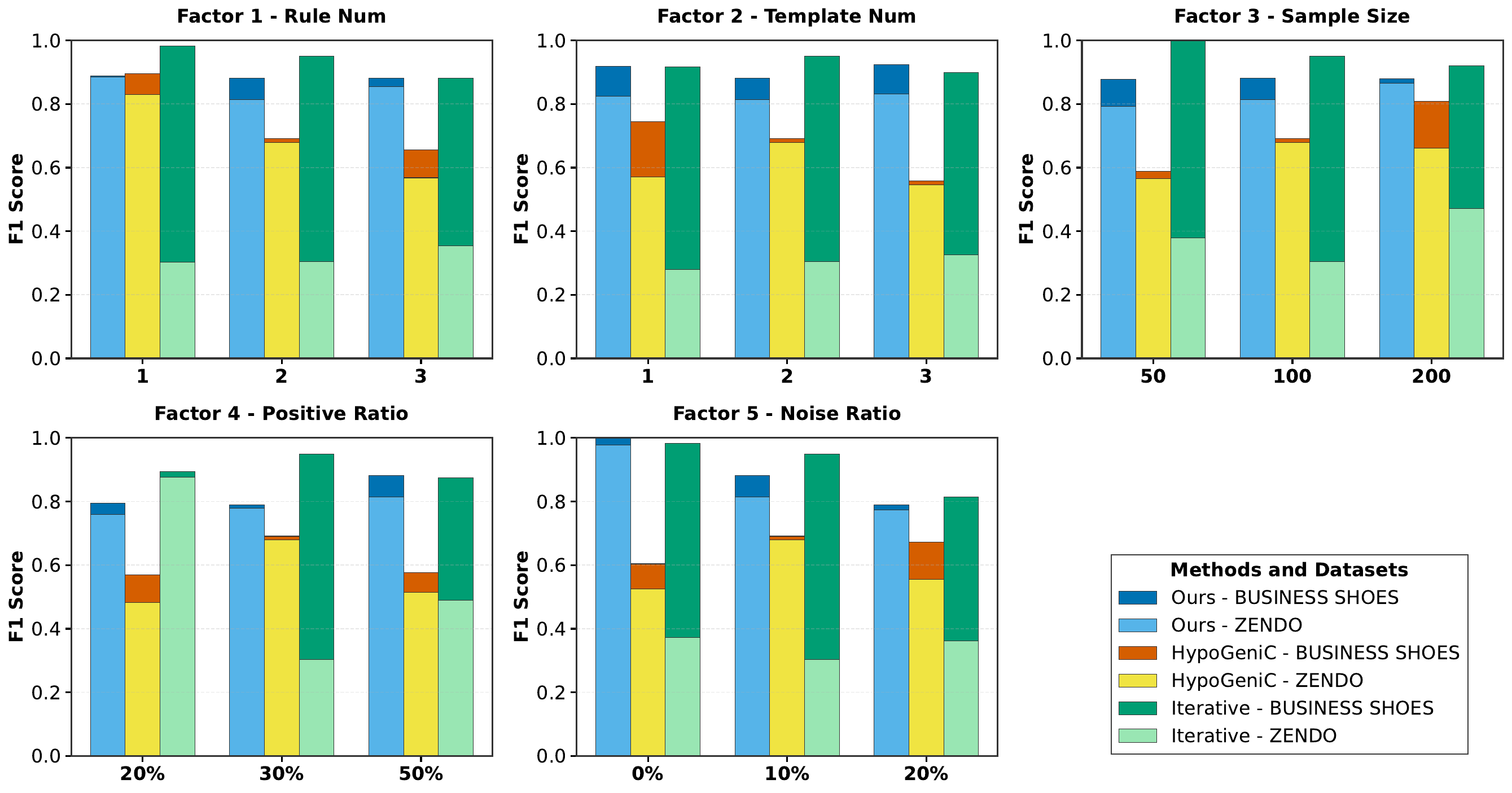}
    \caption{The five subplots present the F1 scores of different methods on the BUSINESS SHOES and ZENDO datasets, each examining one key experimental variable: rule number, template number, sample size, positive ratio, and noise ratio.}
    \label{fig:sub_exp}
\end{figure*}

\subsubsection{Visualization of Results}
Table \ref{tab:case_study} illustrates how the three methods approach hypothesis generation differently on the same Zendo example. IHR generates code to verify hypotheses but achieves only 60\% accuracy, as it struggles with complex relational reasoning despite its structured JSON input processing. HypoGeniC processes natural language descriptions directly but incorrectly infers a rule about "green pieces" (55\% accuracy), demonstrating how pure LLM approaches can form plausible but incorrect generalizations. Our method, using the same natural language input as HypoGeniC, combines LLM capabilities with symbolic ILP solving to correctly identify the position-based pattern, achieving significantly higher accuracy. This example highlights how our hybrid approach effectively overcomes the limitations of both pure neural and symbolic reasoning techniques.

\subsection{Ablation Studies on Data-level Variables}

\textbf{Impact of rule complexity.} As the number of underlying rules increases from 1 to 3, all methods experience performance degradation, but to varying degrees. Our method demonstrates the greatest stability, maintaining high performance even with increased rule complexity as shown in Figure \ref{fig:sub_exp}. This resilience stems from the systematic logical decomposition provided by our approach, which effectively handles conjunction and disjunction of multiple rules, while baseline methods show more substantial performance drops.

\textbf{Effect of template diversity.} Increasing template diversity has the most pronounced impact on HypoGeniC, which relies entirely on LLMs' natural language understanding capabilities. Our method and IHR show greater robustness to template variations, though through different mechanisms - IHR benefits from partially converting reasoning into code, while our approach gains stability by delegating logical structure identification to the ILP solver.

\textbf{Sensitivity to sample size.} Sample size experiments reveal that our method achieves optimal performance even with relatively small datasets, benefiting from the ILP solver's ability to identify the most consistent hypothesis with limited examples. With just 50 samples, our method already achieves performance comparable to what baselines reach with twice as many examples, as demonstrated in Figure \ref{fig:sub_exp}.

\textbf{Robustness to class imbalance.} Varying the positive ratio from 20\% to 50\% demonstrates another key advantage of our approach. While all methods perform better with balanced datasets, our method maintains consistently strong performance even with significant class imbalance. The performance gap between our method and the baselines widens as the positive ratio decreases.

\textbf{Resilience to label noise.} As noise increases from 0\% to 20\% shown in Figure \ref{fig:sub_exp}, both our method and IHR show performance degradation, which is expected since rule-based approaches cannot perfectly represent noisy samples even with ground truth rules. HypoGeniC's performance remains relatively stable across noise levels, reflecting its dependence on the LLM's inherent capabilities rather than strict rule adherence. Nevertheless, despite this degradation, our method and IHR still outperform HypoGeniC in high-noise scenarios, demonstrating the fundamental advantage of structured approaches when dealing with complex reasoning tasks, even under challenging conditions.

\begin{table*}[!t]
    \centering
    \begin{tabular}{l p{5cm} p{5cm} c}
    \toprule
    \textbf{Method} & \textbf{Input} & \textbf{Output} & \textbf{Acc} \\
    \midrule
    IHR & \textcolor{red}{\textbullet} Structured example: 
    \begin{lstlisting}[basicstyle=\ttfamily\tiny, frame=none, aboveskip=2pt, belowskip=2pt]
{
  "pieces": {
    "p36_0": {
      "orientation": "lhs", 
      "size": "small", 
      "color": "green", 
      "position": [4, 5]
    }, 
    "p36_1": {
      "orientation": "strange", 
      "size": "small", 
      "color": "red", 
      "position": [2, 4]
    }, 
    "p36_2": {
      "orientation": "lhs", 
      "size": "large", 
      "color": "blue", 
      "position": [4, 6]
    }, 
    "p36_3": {
      "orientation": "strange", 
      "size": "medium", 
      "color": "blue", 
      "position": [0, 3]
    }
  }, 
  "contacts": []
}
    \end{lstlisting} & 
    \textcolor{red}{\textbullet} Code example: 
    \begin{lstlisting}[basicstyle=\ttfamily\tiny, frame=none, aboveskip=2pt, belowskip=2pt]
def fn(x):
    pieces = x["pieces"]
    
    # Check if any pieces are in contact
    if x["contacts"]:
        return "No"
    
    # Get y-coordinates of all pieces
    y_coords = [piece["position"][1] for piece_id, piece in pieces.items()]
    
    # Check if all y-coordinates are even or all are odd
    all_even = all(y % 2 == 0 for y in y_coords)
    all_odd = all(y % 2 == 1 for y in y_coords)
    
    if all_even or all_odd:
        return "Yes"
    else:
        return "No"
    \end{lstlisting} & 60.0 \\
    \cmidrule(lr){1-4}
    HypoGeniC & \textcolor{red}{\textbullet} Text example: \newline "World 36: piece p36\_0 is lhs-oriented, small and green, located at (4,5); piece p36\_1 is strange-oriented, small and red, located at (2,4); piece p36\_2 is lhs-oriented, large and blue, located at (4,6); piece p36\_3 is strange-oriented, medium and blue, located at (0,3). World 36 is Zendo" & \textcolor{red}{\textbullet} Text example: \newline "The world satisfies the Zendo rule if it contains at least two green pieces." & 55.0 \\
    \cmidrule(lr){1-4}
    Ours & \textcolor{red}{\textbullet} Same with HypoGeniC & \textcolor{red}{\textbullet} Logic program: 
    \begin{lstlisting}[basicstyle=\ttfamily\tiny, frame=none, aboveskip=2pt, belowskip=2pt]
zendo_world(A):- has_piece(A,C), contacts(C,B), blue(B).
zendo_world(A):- has_piece(A,B), strange_oriented(B), large(B).
    \end{lstlisting} & 85.0 \\
    \bottomrule
\end{tabular}

      \caption{
      Case study comparing the inputs, outputs, and performance (Acc, \%) and  of different hypothesis generation methods on the Zendo dataset. The example shows how each method processes the same input differently, with IHR employing code-based verification, HypoGeniC using natural language processing, and our method combining LLM capabilities with symbolic reasoning to achieve higher accuracy.
      }
    \label{tab:case_study}
\end{table*}

\subsection{Overall Analysis}

Our comprehensive results demonstrate that our method consistently outperforms existing approaches across all evaluated dimensions. The key strength lies in our principled task decomposition: leveraging LLMs for natural language understanding and information extraction while delegating logical consistency reasoning to symbolic ILP solvers. This complementary design creates a synergistic system that maintains robust performance across varied tasks, models, and challenging data conditions, proving more effective than approaches relying predominantly on either LLMs or symbolic methods alone.

\section{Conclusion}
\paragraph{Contribution.}
This study proposes an innovative multi-agent collaborative reasoning framework for automatic hypothesis generation and verification.  Our approach effectively reduces the reliance of traditional ILP on expert-defined language bias, achieving an end-to-end automated pipeline from unstructured textual data to verifiable logical hypotheses. Extensive experiments on diverse datasets and challenging scenarios demonstrate that our framework consistently outperforms existing baselines under various data conditions, exhibiting superior performance and robustness. This work not only extends the application of ILP to unstructured text domains, but also provides a new paradigm for building interpretable hybrid AI reasoning systems, laying a solid foundation for automated knowledge discovery.

\paragraph{Limitation.}
Although our method demonstrates its effectiveness in the current experimental settings—including the synthetic SHOES dataset and the Zendo cognitive reasoning task—its performance and applicability to more complex and diverse real-world data (e.g., richer textual content with highly sparse information or more ambiguous semantics) remain to be further explored and validated.


\clearpage
\bibliography{aaai2026}      

\clearpage
\appendix
\onecolumn
\section{MAXSYNTH}
The core idea of the MAXSYNTH algorithm is to introduce the Minimum Description Length (MDL) principle into the search and optimization process of Inductive Logic Programming (ILP). Specifically, MDL advocates selecting the model with the lowest "description cost" when balancing hypotheses and data: on one hand, it seeks to use more compact rule expressions (reducing the complexity of the hypothesis), while on the other hand, it aims to minimize misclassification of observed data (reducing the data residual). In MAXSYNTH, this principle is quantified by computing the total cost of "program size + number of false positives + number of false negatives" for candidate logic programs. The program with the lowest total cost is selected as the output hypothesis, ensuring a balance between model simplicity and good data coverage, even in high-noise environments.

Unlike previous ILP algorithms, MAXSYNTH does not strictly pursue "full coverage of positive examples and zero coverage of negative examples." Instead, it allows a certain degree of error in order to ensure that a suitable explanation can still be found in complex or high-noise data scenarios. Additionally, compared to traditional greedy rule learning or closed searches based on manually defined predicate libraries, MAXSYNTH features a more flexible predicate invention mechanism and the ability to learn recursive rules. It adopts an iterative "generation–combination–constraint" process, integrating an optimal solver based on MaxSAT to progressively eliminate candidate programs that do not satisfy the optimal MDL criterion. This ensures that the remaining programs are globally optimal or near-optimal in terms of both size and misclassification cost. The process is theoretically safeguarded by a "noise-tolerant" constraint that prevents the elimination of any hypothesis that could potentially be an MDL-optimal solution, thereby guaranteeing the correctness and completeness of the algorithm.

\section{Datasets Details}
\label{sec:appendix}
\subsection{BUSINESS SHOES}
\label{sec:business-shoes}

During the construction of the BUSINESS SHOES dataset, we implemented flexible design choices in three key aspects to ensure that our algorithm could be tested for robustness and generalization across various scenarios.

\paragraph{(1) Feature System}  
Each shoe is characterized by five major attributes: \textit{color} (\{\texttt{red}, \texttt{blue}, \texttt{black}, \texttt{white}, \texttt{gray}\}), \textit{material} (\{\texttt{leather}, \texttt{canvas}, \texttt{mesh}, \texttt{synthetic leather}\}), \textit{style} (\{\texttt{sneakers}, \texttt{casual shoes}, \texttt{formal shoes}, \texttt{skateboard shoes}\}), \textit{price} (\{\texttt{cheap}, \texttt{moderate}, \texttt{expensive}\}), and \textit{comfort} (\{\texttt{very comfortable}, \texttt{fairly comfortable}, \texttt{moderately comfortable}\}). This multi-dimensional feature combination provides a rich attribute space for subsequent logical induction.

\paragraph{(2) Variable Natural Language Templates}  
To generate diverse textual descriptions, we provide three different natural language templates for constructing shoe descriptions, as follows:
\begin{itemize}
    \item \texttt{This is a \{\$color\} \{\$style\} made of \{\$material\}, \{\$price\} in price and \{\$comfort\} to wear. This shoe is \{\$conclusion\}.}
    \item \texttt{This \{\$style\} is made of \{\$material\}, comes in \{\$color\}, positioned at a \{\$price\} price point, and is \{\$comfort\}. It is \{\$conclusion\}.}
    \item \texttt{A \{\$color\} \{\$material\} \{\$style\}, priced \{\$price\}, and \{\$comfort\} when worn. The shoe is \{\$conclusion\}.}
\end{itemize}

In our experiments, the number of templates used is controlled by a hyperparameter $N$, which can take values from \{1,2,3\}. When $N > 1$, each sample randomly selects one of the $N$ templates for generation, simulating diverse expression styles.

\paragraph{(3) Variable Decision Rules}
The BUSINESS SHOES dataset defines three default rules for determining whether a shoe is considered “suitable for business occasions” (i.e., a positive example): \begin{itemize} \item \textbf{Formal Business Occasions}: The shoe must satisfy \textit{material} = \texttt{leather}, \textit{color} = \texttt{black}, \textit{style} = \texttt{formal shoes}, and \textit{price} = \texttt{expensive}. \item \textbf{Business Casual Occasions}: The shoe must satisfy \textit{material} = \texttt{synthetic leather}, \textit{style} = \texttt{casual shoes}, and \textit{comfort} = \texttt{very comfortable}. \item \textbf{Modern Formal Business Occasions}: The shoe must satisfy \textit{material} = \texttt{leather}, \textit{style} = \texttt{formal shoes}, with \textit{color} chosen from {\texttt{white}, \texttt{blue}} and \textit{price} = \texttt{moderate}, as well as \textit{comfort} = \texttt{very comfortable}. \end{itemize}

A sample is labeled as positive if it satisfies any one of the three rules; otherwise, it is labeled as negative. In practical experiments, we may choose to use only the first $M$ rules (e.g., using only one, two, or all three rules) to evaluate the model’s performance under different rule complexities. Similar to variable templates, this flexible rule design enables more systematic and comprehensive testing of the algorithm.

In summary, the BUSINESS SHOES dataset serves as a flexible and diverse experimental platform for open-domain hypothesis generation and inductive logic programming tasks by incorporating multi-dimensional features, variable natural language templates, and adjustable decision rules.

\subsection{ZENDO}  
The ZENDO dataset originates from a multi-object logical reasoning scenario, primarily designed to evaluate a model’s ability to perform inductive learning on spatial relationships, object interactions, and attribute compositions. Compared to BUSINESS SHOES, which contains only a single object per sample, each ZENDO sample typically consists of multiple objects (\texttt{piece}), which may have spatial relationships such as contact (\texttt{contact}) or shared coordinates, making the reasoning task more challenging.

\paragraph{(1) Feature System}  
In each \textit{world}, multiple objects are randomly generated, each possessing the following core attributes:  
\begin{itemize}
    \item \textbf{Position coordinates (x, y)}: Defines the object's location in a 2D plane, which determines spatial distribution and potential interactions between objects.
    \item \textbf{Size (size)}: Initially represented as a numerical value, later mapped to abstract categories such as \{\texttt{small}, \texttt{medium}, \texttt{large}\}.
    \item \textbf{Color (color)}: Includes options such as \texttt{red}, \texttt{blue}, and \texttt{green}. Some logical rules may specify particular color combinations.
    \item \textbf{Orientation (orientation)}: Possible values include \texttt{lhs}, \texttt{rhs}, and \texttt{upright}, indicating the object's spatial orientation in the world.
\end{itemize}
Additionally, if two objects are sufficiently close or have adjacent coordinates, they are marked as being in \texttt{contact} during dataset generation. In the implementation, the $i$-th sample’s $j$-th object is uniquely identified using the format \texttt{p\_i\_j}.

\paragraph{(2) Variable Natural Language Templates}  
To generate the natural language description of each ZENDO sample, we randomly select one of the following templates for each object:  
\begin{itemize}
    \item \texttt{\small piece \{\$id\} is a \{\$size\} \{\$color\} piece at (\{\$x\},\{\$y\}) oriented \{\$orientation\}}
    \item \texttt{\small piece \{\$id\} is \{\$orientation\}-oriented, \{\$size\} and \{\$color\}, located at (\{\$x\},\{\$y\})}
    \item \dots
\end{itemize}
These templates highlight different attributes and relationships (e.g., position, orientation, and color), enriching the variety and realism of the textual descriptions. The descriptions of all objects in a world are concatenated, and if any objects are in contact (\texttt{contact}), this relationship is explicitly stated as \texttt{"piece \{\$id1\} contacts piece \{\$id2\}"}. Finally, a summary statement such as \texttt{"World \{\$world\_id\} is/is not Zendo"} is appended.

\paragraph{(3) Variable Decision Rules}  
The dataset defines three configurations, \texttt{Zendo1}, \texttt{Zendo2}, and \texttt{Zendo3}, each corresponding to a scenario with 1, 2, or 3 logical rules:

\begin{itemize}
    \item \textbf{Zendo1}:  
    \begin{center}
    \texttt{zendo1(A) :- piece(A,C), size(C,B), blue(C), small(B), contact(C,D), red(D).}
    \end{center}
    This rule states that a world \texttt{A} satisfies \texttt{zendo1} if there exists an object \texttt{C} that is \texttt{blue}, \texttt{small}, and in \texttt{contact} with a \texttt{red} object \texttt{D}.
    
    \item \textbf{Zendo2}:  
    This setting includes two rules, often involving more complex constraints on colors and coordinates:  
    \begin{center}
    \texttt{zendo2(A) :- piece(A,B), piece(A,C), piece(A,D), green(D), red(B), blue(C).}  
    \texttt{zendo2(A) :- piece(A,B), coord1(B,C), green(D), lhs(B), coord1(D,C).}
    \end{center}
    
    \item \textbf{Zendo3}:  
    This setting extends to three rules, potentially involving conditions on colors, sizes, orientations, and contact relationships:  
    \begin{center}
    \texttt{zendo3(A) :- piece(A,D), blue(D), coord1(D,B), piece(A,C), coord1(C,B), red(C).}\\
    \texttt{zendo3(A) :- piece(A,D), contact(D,C), rhs(D), size(C,B), large(B).}\\
    \texttt{zendo3(A) :- piece(A,B), upright(B), contact(B,D), blue(D), size(D,C), large(C).}
    \end{center}
\end{itemize}

A world is classified as "Zendo" if it satisfies any of the defined logical rules; otherwise, it is classified as "Not Zendo."  

The dataset thus spans multiple levels of complexity, from small-scale (\texttt{zendo1}) to medium (\texttt{zendo2}) and high complexity (\texttt{zendo3}), posing increasing challenges for evaluating a model’s inductive reasoning capabilities and robustness.

\end{document}